\documentclass[10pt,twocolumn,letterpaper]{article}

\usepackage[pagenumbers]{cvpr} 

\usepackage{algorithm}
\usepackage{algorithmic}
\usepackage{amsmath}
\usepackage{amsthm}
\usepackage{makecell}
\usepackage{multirow}
\usepackage{amssymb}

\usepackage{booktabs}
\usepackage{makecell}
\usepackage{graphicx}
\usepackage{bm}
\usepackage{subcaption}

\usepackage[dvipsnames]{xcolor}

\definecolor{cvprblue}{rgb}{0.21,0.49,0.74}
\usepackage[pagebackref,breaklinks,colorlinks,citecolor=cvprblue]{hyperref}

\begin{document}
\title{Improving Adversarial Transferability by Stable Diffusion}

\author{Jiayang Liu$ ^{1} $ \qquad
Siyu Zhu$ ^{2} $ \qquad
Siyuan Liang$ ^{1} $ \qquad
Jie Zhang$ ^{3} $ \qquad
\\
Han Fang$ ^{1} $ \qquad
Weiming Zhang$ ^{4} $ \qquad
Ee-Chien Chang$ ^{1} $ 
\\
$ ^{1} $National University of Singapore
\\
$ ^{2} $Shanghai Jiao Tong University
\\
$ ^{3} $Nanyang Technological University
\\
$ ^{4} $University of Science and Technology of China
\\
{\tt\small \{ljyljy, siyuan96, fanghan\}@nus.edu.sg, siyu\_zhu\_2020@sjtu.edu.cn}
\\
{\tt\small jie\_zhang@ntu.edu.sg, zhangwm@ustc.edu.cn, changec@comp.nus.edu.sg}
}
\maketitle
\begin{abstract}

Deep neural networks (DNNs) are susceptible to adversarial examples, which introduce imperceptible perturbations to benign samples, deceiving DNN predictions. While some attack methods excel in the white-box setting, they often struggle in the black-box scenario, particularly against models fortified with defense mechanisms. Various techniques have emerged to enhance the transferability of adversarial attacks for the black-box scenario. Among these, input transformation-based attacks have demonstrated their effectiveness. In this paper, we explore the potential of leveraging data generated by Stable Diffusion to boost adversarial transferability. This approach draws inspiration from recent research that harnessed synthetic data generated by Stable Diffusion to enhance model generalization.
In particular, previous work has highlighted the correlation between the presence of both real and synthetic data and improved model generalization. Building upon this insight, we introduce a novel attack method called Stable Diffusion Attack Method (SDAM), which incorporates samples generated by Stable Diffusion to augment input images. Furthermore, we propose a fast variant of SDAM to reduce computational overhead while preserving high adversarial transferability. Our extensive experimental results demonstrate that our method outperforms state-of-the-art baselines by a substantial margin. Moreover, our approach is compatible with existing transfer-based attacks to further enhance adversarial transferability.

\end{abstract} 

\section{Introduction}
\label{sec:intro}

In recent years, deep neural networks (DNNs) have brought considerable achievements in various computer vision tasks, including image classification \cite{he2016deep}, object detection \cite{redmon2016you} and face recognition \cite{deng2019arcface}. Nevertheless, prior works \cite{goodfellow2014explaining, szegedy2014intriguing} demonstrated that DNNs are vulnerable to adversarial examples, which add imperceptible perturbations to clean samples to deceive the DNNs. Moreover, adversarial examples often exhibit the property of transferability, which enables those generated for a specific model to effectively mislead other models, thereby exposing threats to DNN models in real-world scenarios. Consequently, exploring methods for generating transferable adversarial examples has attracted great research interest since it can help us better identify the vulnerabilities of neural networks and improve their robustness against adversarial attacks in practical applications.

Various methods have been proposed to improve adversarial transferability, such as advanced optimization algorithm \cite{dong2018boosting, lin2020nesterov, wang2021enhancing}, input transformations \cite{xie2019improving, dong2019evading, lin2020nesterov, zou2020improving} and ensemble-model attacks \cite{liu2017delving, li2020learning}. Among these methods, input transformations have demonstrated effectiveness in boosting adversarial transferability. For example, SIM \cite{lin2020nesterov} computes the average gradient over multiple scale copies of the input image.
Admix \cite{wang2021admix} calculates the gradient of the input image mixed with a small fraction of an additional image randomly sampled from other categories. PAM \cite{zhang2023improving} augments images
from multiple augmentation paths when generating adversarial examples.

Lin \emph{et al}. \cite{lin2020nesterov} regard the generation of adversarial examples on a white-box model as the training process of a neural network, and treat the transferability of adversarial examples as model generalization \cite{lin2020nesterov}. Consequently, the methods utilized to improve model generalization can be extended to the generation of adversarial examples, so as to enhance the transferability of adversarial examples. Recent studies \cite{bansal2023leaving, sariyildiz2023fake} have shown that utilizing synthetic data generated by Stable Diffusion \cite{rombach2022high} during model training can improve model generalization ability. In contrast, we note that existing input transformation-based attacks mainly employ real data for augmentation, which may hinder the transferability of attacks. This encourages us to explore the potential of utilizing synthetic data generated by Stable Diffusion to enhance adversarial transferability.

In this paper, we propose a new attack method named Stable Diffusion Attack Method (SDAM) to improve the transferability of adversarial examples by utilizing the data generated by Stable Diffusion. Addressing the observation made by Bansal \emph{et al}. \cite{bansal2023leaving} that removing either real or synthetic data results in a corresponding reduction in model generalization, we mix the input image with samples generated through Stable Diffusion for augmentation to mitigate this limitation. Furthermore, we propose a fast version of our method to achieve a balance between computational cost and adversarial transferability.

On the other hand, our method can be viewed as augmenting images along a linear path from the mixed image to the pure color image, following the perspective of PAM. While PAM introduces additional augmentation paths to increase the diversity of augmented images, our method explores an alternative way to increase the diversity by mixing up the input image with the samples generated through Stable Diffusion. Moreover, the integration of our method with the PAM strategy can further boost the transferability.

To validate the effectiveness of our method, extensive experiments are conducted against normally trained models, adversarially trained models and defense models. Experimental results show that our method significantly outperforms the state-of-the-art baselines in terms of the attack success rates. Moreover, we evaluate the performance of our method when combined with other transfer-based attacks, and the results demonstrate the compatibility of our method with these attacks.

Our main contributions can be summarized as follows:

\begin{itemize}
    \item We find that existing input transformation-based attacks mainly utilize real data for augmentation, which may limit the transferability of attacks. To address this limitation, we propose a new attack method by leveraging data generated by Stable Diffusion to improve adversarial transferability.
    
    \item We propose a new input transformation by mixing up the input image with data generated by Stable Diffusion. To reduce the computation overhead while maintaining high adversarial transferability, the fast version of our method is also proposed.

    \item Extensive experiments are conducted to validate the effectiveness of our method. Experimental results confirm that our method outperforms the state-of-the-art baselines by a substantial margin. Besides, our method is compatible with other transfer-based attacks to further improve the adversarial transferability.
\end{itemize}

\begin{figure*}[t]
	\centering\includegraphics[width=0.77 \textwidth]{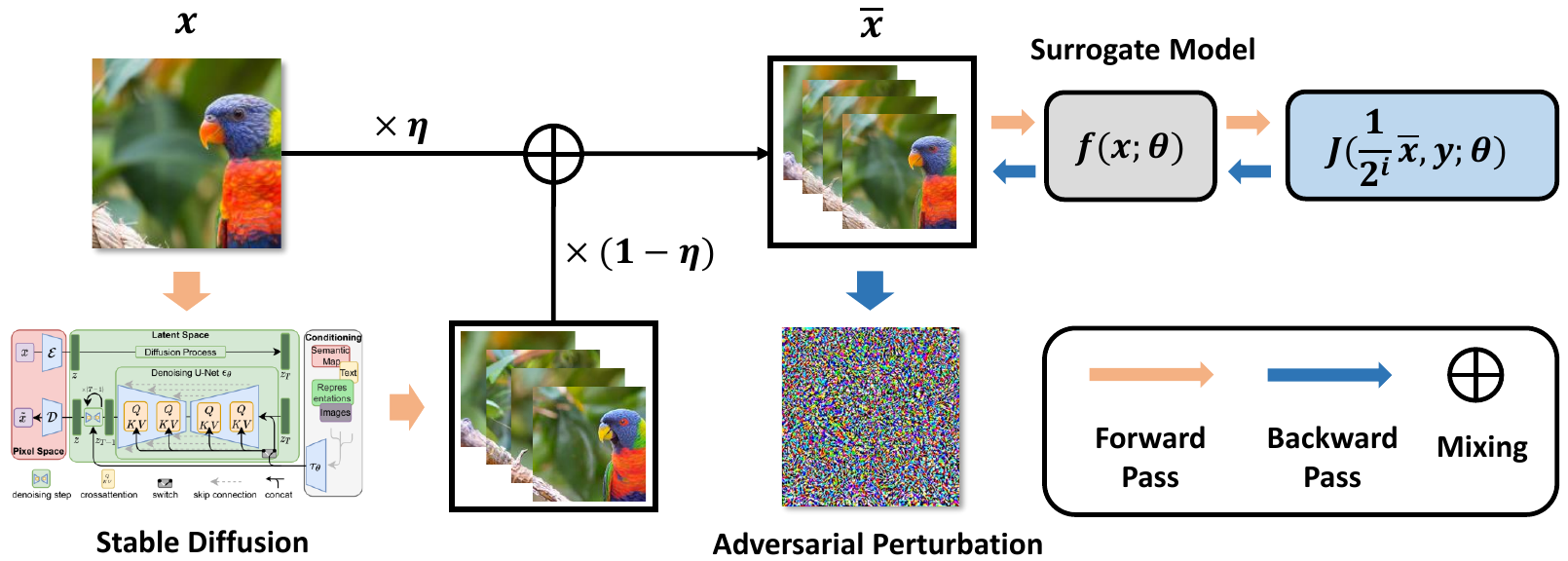}
	\caption{The overall framework of Stable Diffusion Attack Method.}
	\label{fig:framwork}
\end{figure*}

\section{Related Works}
\subsection{Adversarial Attacks}
Adversarial attacks are generally divided into two categories: white-box attacks and black-box attacks. In the white-box attacks, attackers have full knowledge of the victim model. For example, Fast Gradient Sign Method (FGSM) \cite{goodfellow2014explaining} generates adversarial examples by introducing perturbations in the direction of the gradient with a single step. BIM \cite{kurakin2016adversarialmachine} is an iterative approach that applies the FGSM repeatedly in multiple iterations.
In the black-box attacks, attackers have no or limited knowledge about the victim model. 

There are two categories of black-box attacks: query-based attacks and transfer-based attacks. Query-based attacks iteratively query the victim model to obtain gradient information and subsequently optimize the input to deceive the model's predictions. However, the practical application of query-based attacks may be hindered by the potentially prohibitive query costs. Transfer-based attacks generate adversarial examples on a surrogate model and utilize their effectiveness to deceive the victim model. Various methods are proposed to improve adversarial transferability, such as advanced optimization algorithms, input transformations and ensemble-model attacks. For example, MIM \cite{dong2018boosting} is an enhanced version of BIM by adding momentum to the iterative process of perturbing the input data. DIM \cite{xie2019improving} boosts adversarial transferability by applying random resize and padding transformation to the input. TIM \cite{dong2019evading} generates transferable adversarial examples by calculating average gradients from multiple translated images. SIM \cite{lin2020nesterov} utilizes the scale invariance property of DNNs and computes the average gradient over different scaled inputs. Admix \cite{wang2021admix} computes the gradient of the input image mixed with a small fraction of an additional image which is randomly sampled from different categories. PAM \cite{zhang2023improving} augments images
from multiple augmentation paths when generating adversarial examples.

\subsection{Adversarial Defenses}

Various adversarial defenses have been proposed to mitigate the impact of adversarial examples. Adversarial training \cite{goodfellow2014explaining, kurakin2016adversarialmachine, madry2018towards} is one of the most effective methods, which augments the training data with adversarial examples during the training process to make the trained models robust against adversarial attacks. However, adversarial training takes high training cost, especially when dealing with large-scale datasets.
On the other hand, pre-processing based methods are proposed by researchers to defend against adversarial attacks. Liao \emph{et al}. \cite{liao2018defense} propose high-level representation guided denoiser (HGD) to mitigate adversarial examples. Xie \emph{et al}. \cite{xie2018mitigating} perform random resizing and padding on the input images to eliminate adversarial perturbations. Jia \emph{et al}. \cite{jia2019comdefend} propose an end-to-end image compression model to mitigate adversarial examples. Xu \emph{et al} \cite{xu2018feature} investigate the bit depth squeezing to mitigate the adversarial perturbations of the input. Naseer \emph{et al}. \cite{naseer2020self} adversarially train a neural representation purifier (NRP) to remove the adversarial perturbations of the input images. Additionally, there are certified defenses that have been proposed to offer a verifiable robustness guarantee within a defined radius, such as randomized smoothing \cite{cohen2019certified}. In this paper, we utilize these state-of-the-art defenses to evaluate the effectiveness of our method against defense models.

\subsection{Diffusion Models}

Denoising diffusion probabilistic models (DDPMs) \cite{ho2020denoising} are a class of generative models that perform an iterative image denoising process on a random Gaussian noise initial. A U-Net \cite{ronneberger2015u} like neural network is trained based on the minimization of the following objective function:
\begin{equation}
  \mathcal{L} = \mathbb{E}_{\mathbf{x}_0,\bm{\epsilon}\sim N(\mathbf{0},\mathbf{I}),t} \Vert \bm{\epsilon}-\bm{\epsilon}_\theta(\mathbf{x}_t,t)\Vert_2^2
\end{equation}
where $\bm{\epsilon}_\theta$ is the network for predicting the noise $\bm{\epsilon}$ that is applied on the input clean image $\mathbf{x}_0$ with different intensities: $\mathbf{x}_t = \sqrt{\overline{\alpha}_t} \mathbf{x}_0 +  \sqrt{1 - \overline{\alpha}_t} \bm{\epsilon}$, where $\overline{\alpha}_t$ are a series of fixed hyper-parameters. By gradually denoising for $T$ timesteps, the diffusion models can then generate images following the real data distribution from a randomly sampled noise image. These models can further generalize to conditional generations, guided by image classes or texts, and the predicted noise changes into $\bm{\epsilon}_\theta(\mathbf{x}_t,t,\mathcal{C})$, where $\mathcal{C}$ denotes the additional guidance.

As the denoising process operates repeatedly on the high-dimensional image space, training and inference of diffusion models become extremely expensive. To mitigate this, Rombach \emph{et al}. \cite{rombach2022high} propose an approach where the input image is first mapped to a latent space using an autoencoder network prior to the forward and reverse diffusion processes. This method incorporates both self-attention and cross-attention mechanisms \cite{vaswani2017attention} within the network, enhancing guidance from various modalities. By training on large amounts of image-text data pairs \cite{schuhmann2022laion}, this approach is later developed into the well-known Stable Diffusion, which is open source and illustrates excellent generative performance.

\section{Proposed Method}
\subsection{Preliminaries}

Let $\mathbf{x}$ be a clean image and $y$ be the true label for $\mathbf{x}$. Let $l(\mathbf{x})$ be the logits of the classifier $f$ and $J(l(\mathbf{x}),y)$ be the loss function of the classifier ($J$ is often the cross-entropy loss). For brevity, we abbreviate $J(l(\mathbf{x}),y)$ to $J(\mathbf{x},y)$ when there is no ambiguity. The objective of an adversarial attack is generating an adversarial example $\mathbf{x}^{adv}$ that satisfies $\left\|\mathbf{x}-\mathbf{x}^{a d v}\right\|_{p}<\epsilon$ and makes the classifier output an incorrect prediction, where $\|\cdot\|_{p}$ denotes $L_p$-norm distance. In this paper, we focus on $L_{\infty}$-norm of adversarial perturbation to align with previous transfer-based attacks \cite{dong2018boosting, lin2020nesterov}.

\textbf{SIM} \cite{lin2020nesterov} introduces the scale invariance property of DNNs and calculates the average gradient over multiple scale copies of the input image. This attack is updated as follows: 

\begin{equation} \label{eq2}
\begin{split}
    \bar{\mathbf{g}}_{t+1} & = \frac{1}{m} \sum_{i=0}^{m-1} \nabla_{\mathbf{x}^{adv}_{t}} J( \frac{1}{2^{i}} \cdot  \mathbf{x}^{adv}_{t},y), \\   
    \mathbf{x}^{adv}_{t+1} & = \mathbf{x}^{adv}_t + \alpha \cdot sgn( \bar{\mathbf{g}}_{t+1} ).   
\end{split}
\end{equation}
where $m$ is the number of the scale copies.

\textbf{Admix} \cite{wang2021admix} proposes to compute the gradient of the input image mixed with a small fraction of an added image randomly sampled from other categories. Admix integrates with SIM and the gradient is updated as follows:

\begin{equation} \label{eq3}
\begin{split}
    &\bar{\mathbf{g}}_{t+1}  =  \\
    &\frac{1}{m_1\cdot m_2} \sum_{\mathbf{x}' \in \mathbf{X}'}\sum_{i=0}^{m_1-1} \nabla_{\mathbf{x}^{adv}_{t}} J(\frac{1}{2^{i}} \cdot (\mathbf{x}^{adv}_{t}+\eta \cdot \mathbf{x}' ),y), \\
\end{split}
\end{equation}
where $\eta$ controls the strength of the mixture image and $\mathbf{X}'$ is the set of randomly sampled images from other categories.

\textbf{PAM} \cite{zhang2023improving} views the process of SIM as augmenting images along a linear path and proposes to augment images from multiple augmentation paths to improve the adversarial transferability. PAM calculates the gradient as follows:

\begin{equation} \label{eq4}
\begin{split}
    \bar{\mathbf{g}}_{t+1} & = \frac{1}{m} \sum_{i=0}^{m-1} \nabla_{\mathbf{x}^{adv}_{t}} J(\frac{1}{2^i} \cdot \mathbf{x}_t^{adv} + (1-\frac{1}{2^i}) \cdot \mathbf{x}',y) \\
\end{split}
\end{equation}
where $\mathbf{x}'$ is the baseline image from the path pool.

\begin{table*}[htbp]
\caption{Attack success rates (\%) on normally trained models and adversarially trained models by various transfer-based attacks. The best results are marked in bold.}
\centering
\scalebox{0.9}{
\begin{tabular}{@{}ccccccccc@{}}
\toprule[1.5pt]
\multirow{2}{*}{\begin{tabular}[c]{@{}c@{}}Surrogate\\Model\end{tabular}} & \multirow{2}{*}{Attack} & \multicolumn{7}{c}{Models} \\ \cmidrule(lr){3-9}

 &  &Inc-v3  &Inc-v4 &IncRes-v2 &Res-101 &Inc-v3$_{ens3}$  &Inc-v3$_{ens4}$  &IncRes-v2$_{ens}$  \\ \midrule[1.2pt]

\multirow{6}{*}{Inc-v3} 
 & DIM &95.9	&62.2	&58.5	&51.3	&22.2	&21.6	&11.1  \\
 & TIM &97.9	&46.8	&40.9	&37.7	&25.6	&24.7	&16.2   \\
 
 & SIM &97.9	&68.0	&65.5	&60.4	&35.3	&35.7	&19.7   \\
 & Admix &\textbf{98.4}	&77.1	&73.6	&67.4	&38.3	&37.1	&22.0  \\
 & PAM &98.3	&80.2	&77.3	&73.4	&45.6	&44.6	&26.2   \\ \cmidrule(l){2-9} 
 & SDAM-Fast &98.0	&90.2	&90.0	&86.6	&60.2	&61.0	&37.0   \\ 
 & SDAM &98.1	&\textbf{93.7}	&\textbf{93.1}	&\textbf{90.9}	&\textbf{72.3}	&\textbf{71.4}	&\textbf{46.8}   \\ \midrule[1.2pt]
\multirow{6}{*}{Inc-v4} 
 & DIM &76.0	&98.2	&65.2	&55.8	&26.8	&23.6	&14.5   \\
 & TIM &60.9	&98.7	&46.0	&39.6	&26.7	&26.8	&20.4   \\
 
 & SIM &84.6	&\textbf{99.3}	&77.0	&69.5	&44.1	&40.9	&28.0   \\
 & Admix &85.9	&98.8	&80.1	&73.0	&49.7	&46.9	&31.7   \\
 & PAM &86.6	&99.1	&80.3	&75.8	&54.1	&53.3	&35.3   \\ \cmidrule(l){2-9} 
 & SDAM-Fast &93.8	&98.3	&91.0	&89.7	&75.2	&70.9	&54.3   \\ 
 & SDAM &\textbf{95.2}	&98.5	&\textbf{93.5}	&\textbf{91.8}	&\textbf{82.4}	&\textbf{77.9}	&\textbf{62.6}   \\ \midrule[1.2pt]
\multirow{6}{*}{IncRes-v2} 
 & DIM &71.8	&69.6	&96.5	&62.1	&33.6	&30.6	&20.9   \\
 & TIM &64.9	&60.2	&99.0	&49.0	&34.6	&31.9	&27.6   \\
 
 & SIM &86.6	&83.3	&99.8	&76.9	&53.1	&47.7	&37.5 \\
 & Admix &89.0	&86.5	&99.8	&82.7	&62.1	&57.0	&48.3   \\
 & PAM &91.4	&87.7	&\textbf{100.0}	&85.0	&70.9	&65.0	&56.8   \\ \cmidrule(l){2-9} 
 & SDAM-Fast &96.0	&95.1	&99.6	&92.9	&83.0	&75.9	&69.0   \\
 & SDAM &\textbf{96.1}	&\textbf{95.2}	&99.7	&\textbf{93.5}	&\textbf{86.3}	&\textbf{81.0}	&\textbf{76.4}   \\ \midrule[1.2pt]
\multirow{6}{*}{Res-101} 
 & DIM &75.1	&69.8	&69.4	&96.0	&38.8	&36.2	&22.2   \\
 & TIM &59.3	&52.9	&49.6	&97.5	&37.9	&35.7	&26.4   \\
 
 & SIM &78.3	&73.2	&72.7	&\textbf{98.6}	&45.0	&40.4	&26.7   \\
 & Admix &79	&75.6	&73.3	&98.2	&48.9	&45.0	&30.8   \\
 & PAM &80.5	&75.3	&73.8	&98.2	&53.9	&49.3	&35.5   \\ \cmidrule(l){2-9} 
 & SDAM-Fast &90.1	&88.2	&88.4	&97.9	&72.2	&66.6	&50.0  \\ 
 & SDAM &\textbf{92.5}	&\textbf{91.2}	&\textbf{90.1}	&97.6	&\textbf{79.1}	&\textbf{74.6}	&\textbf{60.0}   \\ \bottomrule[1.5pt]

\end{tabular}
}
\label{tab:attacknormally}
\end{table*}

\begin{table*}[htbp]
\caption{The attack success rates (\%) on normally trained models and adversarially trained models in the ensemble-model setting by various transfer-based attacks. The adversarial examples are generated on the ensemble models, including Inc-v3, Inc-v4, IncRes-v2 and Res-101. The best results are marked in bold.}
\centering
\scalebox{0.9}{
\begin{tabular}{@{}cccccccc@{}}
\toprule[1.5pt]
  \multirow{2}{*}{Attack} & \multicolumn{7}{c}{Models} \\ \cmidrule(lr){2-8}

   &Inc-v3  &Inc-v4 &IncRes-v2 &Res-101 &Inc-v3$_{ens3}$  &Inc-v3$_{ens4}$  &IncRes-v2$_{ens}$  \\ \midrule[1.2pt]

  DIM &99.1	&99.1	&98.5	&98.8	&70.3	&64.5	&48.7   \\
  TIM &99.7	&99.7	&99.4	&99.1	&73.0	&68.4	&58.7  \\
  SIM &\textbf{99.9}	&\textbf{99.9}	&99.7	&\textbf{99.6}	&86.5	&83.0	&70.6   \\
  Admix &\textbf{99.9}	&\textbf{99.9}	&99.8	&\textbf{99.6}	&86.5	&83.0	&69.6  \\
  PAM &97.2	&97.3	&97.2	&97.1	&89.0	&86.4	&75.6   \\  \cmidrule(l){1-8}
  SDAM-Fast &\textbf{99.9}	&\textbf{99.9}	&\textbf{99.9}	&\textbf{99.6}	&94.3	&92.9	&85.9   \\
  SDAM &\textbf{99.9}	&\textbf{99.9}	&\textbf{99.9}	&\textbf{99.6}	&\textbf{95.2}	&\textbf{94.4}	&\textbf{89.0}   \\ \bottomrule[1.5pt]

\end{tabular}
}
\label{tab:ens}
\end{table*}

\begin{table*}[htbp]
\caption{The attack success rates (\%) on normally trained models and adversarially trained models setting by various transfer-based attacks combined with augmentation-based strategies. The adversarial examples are generated on the Inception-v3 model. The best results are marked in bold.}
\centering
\scalebox{0.9}{
\begin{tabular}{@{}cccccccc@{}}
\toprule[1.5pt]
  \multirow{2}{*}{Attack} & \multicolumn{7}{c}{Models} \\ \cmidrule(lr){2-8}

   &Inc-v3  &Inc-v4 &IncRes-v2 &Res-101 &Inc-v3$_{ens3}$  &Inc-v3$_{ens4}$  &IncRes-v2$_{ens}$  \\ \midrule[1.2pt]

  SIM-DI &96.1	&83.3	&79.3	&76.1	&46.9	&48.0	&28.8   \\
  Admix-DI &\textbf{98.5}	&89.0	&88.6	&84.1	&53.6	&51.3	&31.3  \\
  PAM-DI &97.3	&89.2	&89.7	&86.4	&63.5	&62.2	&38.3   \\ 
  SDAM-Fast-DI &96.1	&\textbf{92.6}	&\textbf{91.1}	&\textbf{91.0}	&\textbf{77.2}	&\textbf{74.1}	&\textbf{52.6}   \\ \midrule[1.2pt]

  SIM-TI &97.9	&68.8	&64.5	&57.6	&47.8	&47.3	&33.5   \\
  Admix-TI &\textbf{98.6}	&81.0	&76.3	&70.7	&57.7	&57.1	&39.5   \\
  PAM-TI &98.2	&82.7	&78.0	&71.1	&63.8	&61.4	&45.6   \\  
  SDAM-Fast-TI &98.0	&\textbf{91.2}	&\textbf{89.3}	&\textbf{85.2}	&\textbf{79.1}	&\textbf{77.1}	&\textbf{59.1}   \\ \midrule[1.2pt]

  SIM-DI-TI &97.2	&83.5	&78.5	&74.6	&67.5	&63.5	&48.0   \\
  Admix-DI-TI &\textbf{98.8}	&89.1	&86.2	&81.1	&72.4	&70.1	&54.1   \\
  PAM-DI-TI &97.2	&89.8	&87.8	&84.9	&79.3	&77.7	&60.7   \\  
  SDAM-Fast-DI-TI &96.1	&\textbf{92.4}	&\textbf{90.7}	&\textbf{88.3}	&\textbf{86.1}	&\textbf{84.8}	&\textbf{73.0}   \\ \bottomrule[1.5pt]

\end{tabular}
}
\label{tab:di-ti}
\end{table*}

\begin{table*}[htbp]
\caption{The attack success rates (\%) on seven defense models. The adversarial examples are generated on the Inception-v3 model. The best results are marked in bold.}
\centering
\scalebox{0.9}{
\begin{tabular}{@{}ccccccccc@{}}
\toprule[1.5pt]
  \multirow{2}{*}{Attack} & \multicolumn{7}{c}{Defenses} &\multirow{2}{*}{Average} \\ \cmidrule(lr){2-8}

   &HGD  &R\&P &NIPS-r3 &ComDefend &Bit-Red &RS &NRP    \\ \midrule[1.2pt]

  SIM &15.7 &19.1 &25.8 &46.2 &14.9 &26.3 &12.3  &22.9   \\
  Admix &20.6 &22.8 &32.8 &54.6 &22.5 &28.2 &13.4   &27.8 \\
  PAM &25.5 &24.4 &33.1 &56.4 &24.8 &29.9  &14.6 &29.8 \\  \cmidrule(l){1-9}
  SDAM-Fast &36.8 &38.2 &38.2 &70.6 &35.5 &37.3 &17.6  &39.1   \\ 
  SDAM &\textbf{45.7} &\textbf{47.8} &\textbf{60.6} &\textbf{76.8} &\textbf{45.0} &\textbf{41.3} &\textbf{18.0} &\textbf{47.9}    \\ \bottomrule[1.5pt]

\end{tabular}
}
\label{tab:defense}
\end{table*}

\begin{table*}[htbp]
\caption{The attack success rates (\%) on normally trained models and adversarially trained models. The adversarial examples are generated on the Inception-v3 model. The best results are marked in bold.}
\centering
\scalebox{0.9}{
\begin{tabular}{@{}cccccccc@{}}
\toprule[1.5pt]
  \multirow{2}{*}{Attack} & \multicolumn{7}{c}{Models} \\ \cmidrule(lr){2-8}

   &Inc-v3  &Inc-v4 &IncRes-v2 &Res-101 &Inc-v3$_{ens3}$  &Inc-v3$_{ens4}$  &IncRes-v2$_{ens}$  \\ \midrule[1.2pt]

 PAM &\textbf{98.3}	&80.2	&77.3	&73.4	&45.6	&44.6	&26.2   \\ 
 SDAM-Fast &98.0	&90.2	&90.0	&86.6	&60.2	&61.0	&37.0   \\ 
 SDAM-Fast-PAM &98.1	&\textbf{91.4}	&\textbf{90.5}	&\textbf{87.4}	&\textbf{67.3}	&\textbf{65.5}	&\textbf{42.6}   \\  \bottomrule[1.5pt]

\end{tabular}
}
\label{tab:path}
\end{table*}

\subsection{Motivation}
Lin \emph{et al}. \cite{lin2020nesterov} draw an analogy between the generation of adversarial examples and standard neural network training. Thus, we can migrate the methods utilized to boost the generalization of models to the generation of adversarial examples, so as to boost the transferability of adversarial examples. Moreover, Bansal \emph{et al}. \cite{bansal2023leaving} propose that training a classifier on a combination of real data and synthetic data generated by Stable Diffusion can achieve good model generalization ability. On the contrary, we notice that existing input transformation-based attacks mainly utilize real data for augmentation, which may limit the transferability of attacks. This inspires us to leverage data generated by Stable Diffusion  to improve adversarial transferability. 

In this work, we propose a new attack method named Stable Diffusion Attack Method (SDAM) to enhance the transferability of adversarial attack, which mixes the input image with multiple images generated by Stable Diffusion for augmentation.

\subsection{SDAM Method}
Bansal \emph{et al}. \cite{bansal2023leaving} point out that removing either real or generated data results in a corresponding reduction in model generalization. To address this limitation, we mix up the target image with the image generated by Stable Diffusion  $\bar{\mathbf{x}} = \eta \cdot \mathbf{x}_t^{adv} + (1-\eta) \cdot \mathbf{x}_t^{j}$. In our method, the gradient is calculated as follows:

\begin{equation} \label{eq5}
\begin{split}
    &\bar{\mathbf{g}}_{t+1}  =  \\
    &\frac{1}{m\cdot n} \sum_{j=0}^{n-1}\sum_{i=0}^{m-1} \nabla_{\mathbf{x}^{adv}_{t}} J(\frac{1}{2^{i}} \cdot (\eta \cdot \mathbf{x}^{adv}_{t}+(1-\eta) \cdot \mathbf{x}^{j}_{t} ),y), \\
\end{split}
\end{equation}
where $\mathbf{x}^{j}_{t}$ is generated by the pretrained Stable Diffusion model\footnote{https://github.com/CompVis/stable-diffusion} \cite{rombach2022high} whose  input is the target image $\mathbf{x}^{adv}_{t}$ and the prompt is `a photo of $y$' where $y$ is the true label. $n$ is the number of samples generated by Stable Diffusion, $m$ is the number of scale copies and $\eta$ is the mixing ratio which controls the portion of the target image and the image generated by Stable Diffusion. The overall framework of our method is illustrated in Figure \ref{fig:framwork}.

From the perspective of PAM, our method can also be considered as augmenting images along a linear path from the mixed image to the pure color image: $\frac{1}{2^{i}} \cdot  (\eta \cdot \mathbf{x}^{adv}_{t}+(1-\eta) \cdot \mathbf{x}^{j}_{t}) = \frac{1}{2^i} \cdot (\eta \cdot \mathbf{x}^{adv}_{t}+(1-\eta) \cdot \mathbf{x}^{j}_{t}) + (1-\frac{1}{2^i}) \cdot \mathbf{0}$. PAM proposes to explore more augmentation paths to increase the diversity of augmented images. In contrast, we explore another way to increase the diversity by mixing the target image with multiple Stable Diffusion samples. Moreover, these two ways of increasing diversity are compatible. The experimental results in Table \ref{tab:path} illustrate the compatibility of our method with PAM.

We also notice that the computation overhead of the proposed method is large since multiple samples need to be generated by Stable Diffusion in each iteration of the algorithm. Therefore, we propose a fast version of our method (SDAM-Fast) which can significantly reduce the computation overhead while preserving high adversarial transferability. In SDAM-Fast, we only use Stable Diffusion to generate $\mathbf{x}^{j}_{0}$ in the initial iteration. Subsequently, in the following iterations, we repeatedly utilize the $\mathbf{x}^{j}_{0}$ to mix with the target image instead of generating $\mathbf{x}^{j}_{t}$ in each iteration, which significantly reduces the computation overhead. The gradient is computed in SDAM-Fast as follows:

\begin{equation} \label{eq6}
\begin{split}
    &\bar{\mathbf{g}}_{t+1}  =  \\
    &\frac{1}{m\cdot n} \sum_{j=0}^{n-1}\sum_{i=0}^{m-1} \nabla_{\mathbf{x}^{adv}_{t}} J(\frac{1}{2^{i}} \cdot (\eta \cdot \mathbf{x}^{adv}_{t}+(1-\eta) \cdot \mathbf{x}^{j}_{0} ),y), \\
\end{split}
\end{equation}
where $\mathbf{x}^{j}_{0}$ is generated by Stable Diffusion whose input is the initial input image $\mathbf{x}^{adv}_{0}$ and the prompt is still `a photo of $y$' where $y$ is the true label. The fast version of our method makes a balance between the computational cost and adversarial transferability. 

\subsection{Differences with Other Methods}
\begin{itemize}[leftmargin=*,noitemsep,topsep=2pt]
	\item Compared with DIM, TIM, SIM, Admix and PAM, our SDAM method introduces synthetic data generated by Stable Diffusion for improving the transferability of adversarial examples, while these existing methods mainly utilize real data for augmentation.
	\item Our SDAM method preserves some original information by mixing the image with samples generated through Stable Diffusion from the same category, while Admix mixes images with different categories. 
	\item PAM increases the diversity of augmented images by exploring more augmentation paths, while our SDAM method obtains diversity of images by mixing the image with multiple samples generated by Stable Diffusion.
\end{itemize}
\section{Experiments}

\subsection{Experimental Setup}
\subsubsection{Dataset}
We perform experiments on 1,000 images from an ImageNet-compatible dataset\footnote{https://github.com/cleverhans-lab/cleverhans/tree/master/\\cleverhans\_v3.1.0/examples/nips17\_adversarial\_competition} that was used in the NIPS 2017 adversarial competition. This dataset has been widely used in many transferable adversarial attack works \cite{dong2019evading, zou2020improving}.

\subsubsection{Models}
In the experiments, we adopt four normally trained networks as the victim models, including Inception-v3 (Inc-v3) \cite{szegedy2016rethinking}, Inception-v4 (Inc-v4), Inception-Resnet-v2 (IncRes-v2) \cite{szegedy2017inception} and Resnet-v2-101 (Res-101) \cite{he2016deep}. We evaluate our method on three adversarially trained models, namely Inc-v3$_{ens3}$, Inc-v3$_{ens4}$ and IncRes-v2$_{ens}$ \cite{tramer2017ensemble}. We also study seven defense models, including HGD \cite{liao2018defense}, R\&P \cite{xie2018mitigating}, NIPS-r3\footnote{https://github.com/anlthms/nips-2017/tree/master/mmd}, ComDefend \cite{jia2019comdefend}, Bit-Red \cite{xu2018feature}, RS \cite{cohen2019certified} and NRP \cite{naseer2020self}.

\subsubsection{Baselines}
In the experiments, we adopt DIM \cite{xie2019improving}, TIM \cite{dong2019evading}, SIM \cite{lin2020nesterov}, Admix \cite{wang2021admix} and PAM \cite{zhang2023improving} as our baselines. All these methods are integrated into MIM \cite{dong2018boosting}.

\subsubsection{Hyper-parameters}
In the experiments, we set the maximum perturbation of $\epsilon = 16$, the step size $\alpha = 1.6$, the number of iteration $T=10$, and the decay factor $\mu = 1.0$ for all the methods. We set the transformation probability $p = 0.5$ for DIM, the Gaussian kernel with kernel size $7 \times 7$ for TIM, and the number of scaled images $m=5$ for SIM. For Admix, we set $m_1 = 5$, $m_2 = 3$ and $\eta=0.2$. For PAM, we set $m=4$ and the number of augmentation paths is 8. The parameters for these attacks follow the corresponding default settings. For SDAM, we set $\eta=0.6$, $n = 20$, $m = 5$ and the strength of Stable Diffusion is 0.7.

\subsection{Attack a Single Model}
First of all, we compare our method with various input transformation-based attacks, including DIM, TIM, SIM, Admix and PAM, on a single model. We generate the adversarial examples on each normally trained model and test them on all the seven models. Table \ref{tab:attacknormally} shows the attack success rates which are the misclassification rates of the victim models on the generated adversarial examples.

In general, we can see from the results that our method can achieve the best performance of adversarial transferability on black-box models and maintain high performance on white-box models. For instance, SIM, Admix and PAM achieve the attack success rates of 60.4\%, 67.4\% and 73.4\%, respectively on Res-101 when crafting adversarial examples on Inc-v3. In contrast, our method can achieve 90.9\% attack success rate, which is 17.5\% higher than PAM. Moreover, our method achieves the highest adversarial tranferability against adversarially trained models, which indicates the effectiveness of our method. These results confirm our motivation that introducing data generated by Stable Diffusion for augmentation can boost the adversarial transferability, especially for adversarially trained models. In addition, the fast version of our method can also achieve much higher attack success rates than the other baseline methods while reducing the computation overhead.

\subsection{Attack an Ensemble of Models}
It is pointed out by Liu \emph{et al}. \cite{liu2017delving} that attacking several models in parallel can further boost the adversarial transferability. We integrate various adversarial attacks with the ensemble attack method in \cite{dong2018boosting}, which fuses the logit outputs of multiple models. We attack the ensemble of four normally trained models in the experiments, including Inc-v3, Incv4, IncRes-v2 and Res-101. We assign equal weights to all the ensemble models and evaluate the performance of adversarial transferability.

Table \ref{tab:ens} shows the results that our proposed method can always achieve the best attack success rates on the black-box models. Compared with the baseline attacks, our method can achieve the attack success rates of 95.2\%, 94.4\% and 89.0\%, respectively on Inc-v3$_{ens3}$, Inc-v3$_{ens4}$ and IncRes-v2$_{ens}$, which outperforms PAM by more than 6\%. Particularly, the fast version of our method SDAM-Fast can also achieve comparable transferability to SDAM, while reducing the computation overhead.

\subsection{Combined with Input Transformation-based Methods}
Existing input transformation-based attacks exhibit notable compatibility with each other. Similarly, our method can also integrate with other input transformation-based attacks to boost the adversarial transferability. We integrate various attacks with three transformation-based methods, including DIM, TIM and DI-TIM (simultaneously combined with DIM and TIM). Considering the computation overhead, we only integrate the fast version of our method with these transformation-based methods. The adversarial examples are generated on the Inc-v3 model.

The results in Table \ref{tab:di-ti} show that our method achieves the best performance when combined with various transformation-based methods. For example, our method outperforms PAM by 4.6\%, 14.1\% and 3.4\% on Res-101 when combined with DIM, TIM and DI-TIM, respectively. Particularly, our method can achieve much better results on adversarially trained models, which are more robust against adversarial attacks. These results further demonstrate the effectiveness of our method in improving adversarial transferability.

\begin{figure*}[thbp]
  \centering
  \begin{subfigure}{0.246\linewidth}
    \includegraphics[width=\linewidth]{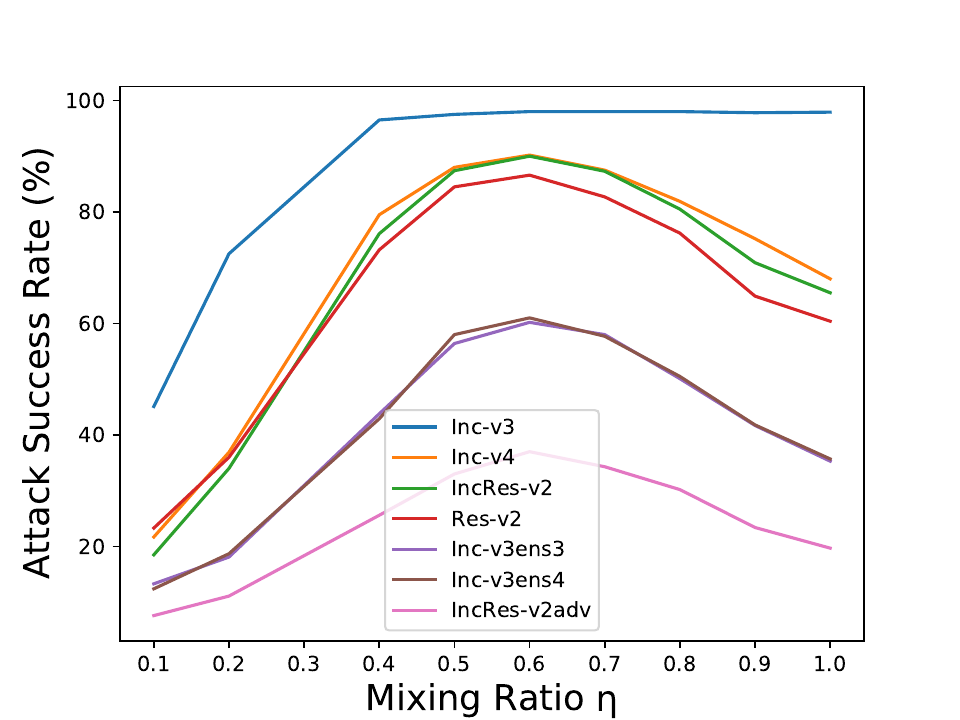}
    \caption{}
    \label{eta}
  \end{subfigure}
  \hfill
  \begin{subfigure}{0.246\linewidth}
    \includegraphics[width=\linewidth]{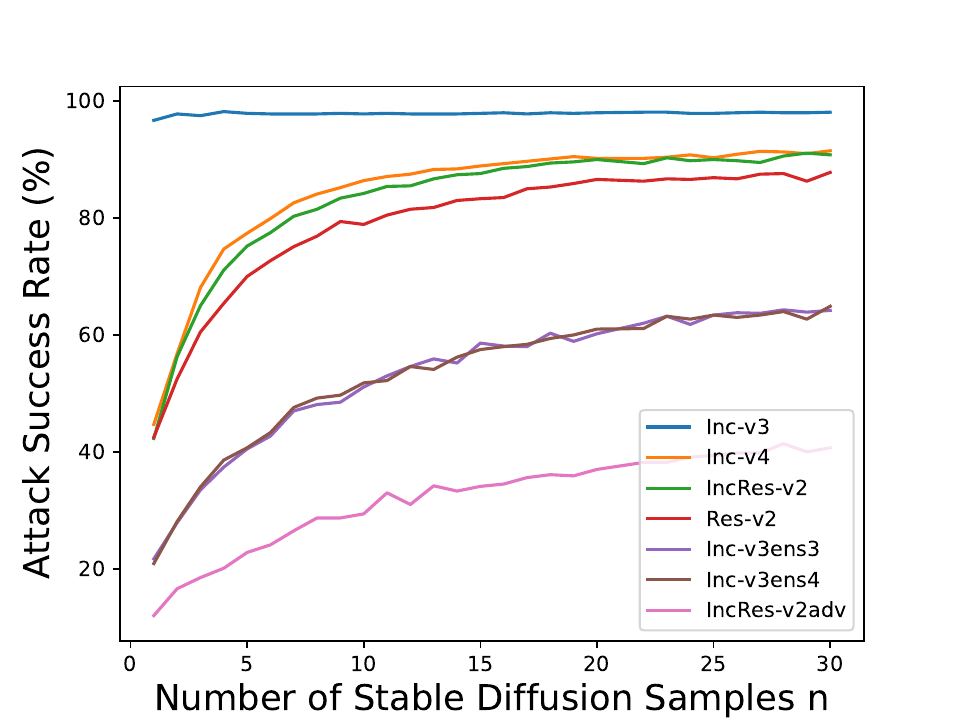}
    \caption{}
    \label{n}
  \end{subfigure}
    \hfill
  \begin{subfigure}{0.246\linewidth}
    \includegraphics[width=\linewidth]{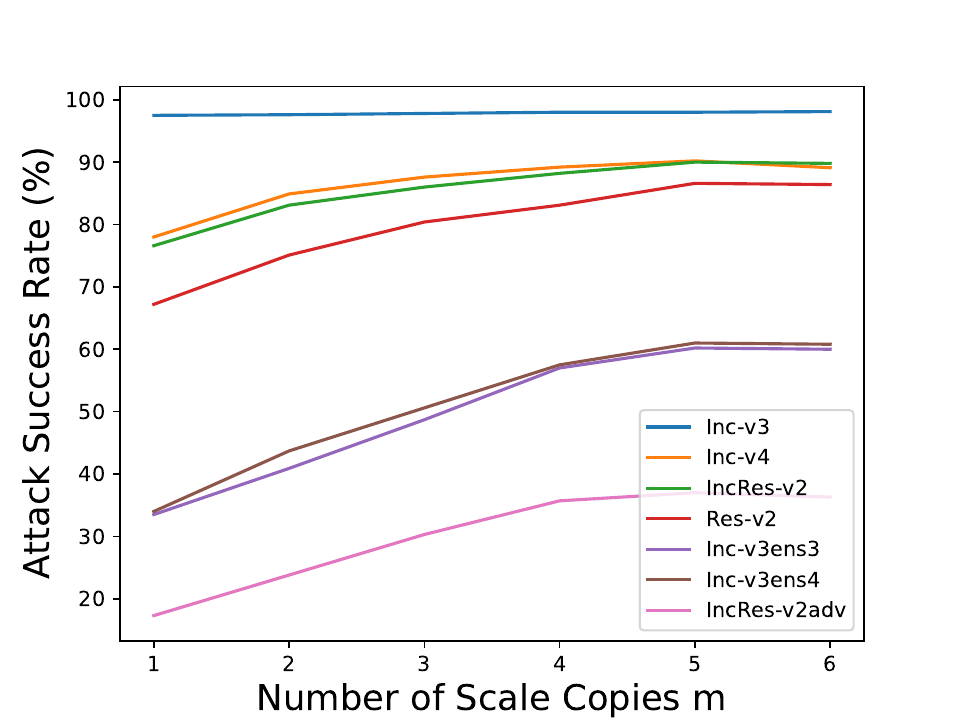}
    \caption{}
    \label{m}
  \end{subfigure}
    \hfill
  \begin{subfigure}{0.246\linewidth}
    \includegraphics[width=\linewidth]{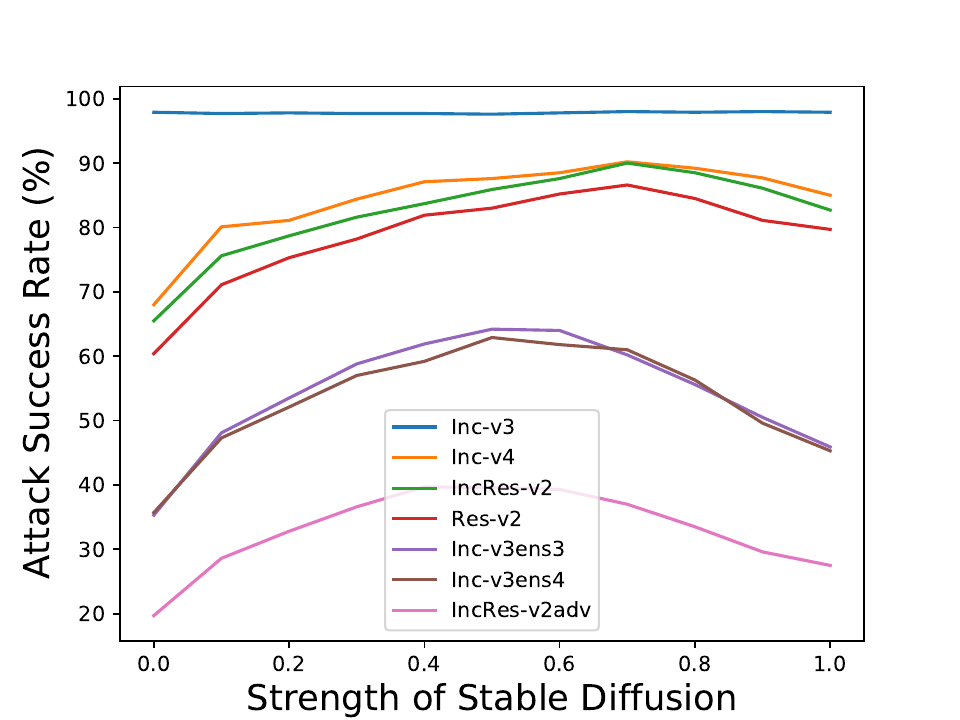}
    \caption{}
    \label{strength}
  \end{subfigure}
  \caption{The attack success rates (\%) of our method with different values of hyper-parameters.}
  \label{ablation}
\end{figure*}

\subsection{Attack Defense Models}
In order to further identify the effectiveness of the proposed method, we evaluate the performance of various adversarial attacks on seven defense methods, including the top-3 defense methods in the NIPS 2017 competition (HGD \cite{liao2018defense}, R\&P \cite{xie2018mitigating} and NIPS-r3), and four widely used defense methods, namely ComDefend \cite{jia2019comdefend}, Bit-Red \cite{xu2018feature}, RS \cite{cohen2019certified} and NRP \cite{naseer2020self}. In the experiments, we use SIM, Admix and PAM as baselines since these three attacks achieve relatively good performance in the above evaluations. We generate the adversarial examples on the Inc-v3 model. The results are reported in Table \ref{tab:defense}.

We can see from Table \ref{tab:defense} that our method still exhibits the best performance against these defenses. For example, SIM, Admix and PAM can only achieve an average attack success rate of 22.9\%, 27.8\% and 29.8\%, respectively. In contrast, our method can achieve an average attack success rate of 47.9\%, which outperforms PAM by a clear margin of 18.1\%. Moreover, the fast version of our method can also achieve an average attack success rate of 39.1\%, which is higher than the other baseline methods while reducing the computation overhead. These results demonstrate that our method is also effective in boosting adversarial transferability against defense models.

\subsection{Combined with Augmentation Path}
Zhang \emph{et al}. \cite{zhang2023improving} proposed that augmenting images from multiple augmentation paths can improve the transferability of the generated adversarial examples. We integrate our method with this strategy and evaluate the performance of adversarial transferability. Considering the computation overhead, we integrate the fast version of our method with an additional augmentation path of PAM, rather than employing all the eight augmentation paths of PAM. The adversarial examples are crafted on the Inc-v3 model.

Table \ref{tab:path} demonstrates that our method integrating with one additional augmentation path can further boost the adversarial transferability. Specially, our method combined with one additional augmentation path achieves better transferability on adversarially trained models, which is 7.1\%, 4.5\% and 5.6\% higher than our method on Inc-v3$_{ens3}$, Inc-v3$_{ens4}$ and IncRes-v2$_{ens}$, respectively. These results illustrate the high compatibility of our method with the existing transfer-based attack strategy.

\subsection{Ablation Studies}
We perform the following ablation studies to investigate the effect of four hyper-parameters: mixing ratio $\eta$, number of Stable Diffusion samples $n$, number of scale copies $m$, and strength of Stable Diffusion. To simplify the analysis, we only consider the transferability of adversarial examples generated on the Inc-v3 by the fast version of our method.

\subsubsection{Mixing Ratio $\eta$}
Figure \ref{eta} illustrates the attack success rates of SDAM-Fast with different values of $\eta$, where $n$, $m$ and $strength$ is fixed to 20, 5 and 0.7, respectively. We can observe from the result that the transferability improves on the other six models when $\eta \leq 0.6$, while the transferability decreases when $\eta > 0.6$. To achieve the best performance of transferability, we choose $\eta=0.6$.

\subsubsection{Number of Stable Diffusion Samples $n$}
Figure \ref{n} illustrates the attack success rates of SDAM-Fast with different values of $n$, where $\eta$, $m$ and $strength$ is fixed to 0.6, 5 and 0.7, respectively. The result shows that the transferability improves on the other six models when $n$ becomes larger. However, a larger value of $n$ leads to higher computation cost. Therefore, we choose $n = 20$ to balance the computational cost and adversarial transferability.

\subsubsection{Number of Scale Copies $m$}
Figure \ref{m} illustrates the attack success rates of SDAM-Fast with different values of $m$, where $\eta$, $n$ and $strength$ is fixed to 0.6, 20 and 0.7, respectively. We can see that the transferability improves on the other six models when $m \leq 5$, while the transferability remains almost unchanged when $m > 5$ . Therefore, we choose $m = 5$ to achieve the best performance of transferability.

\subsubsection{Strength of Stable Diffusion}
Figure \ref{strength} illustrates the attack success rates of SDAM-Fast with different values of $strength$, where $\eta$, $n$ and $m$ is fixed to 0.6, 20 and 5, respectively. We can observe that the transferability improves on most of the models when $strength \leq 0.7$, while the transferability decreases on all the other six models when $strength > 0.7$. Therefore, we choose $strength=0.7$ to achieve the best performance of transferability.

\section{Conclusion}
Inspired by recent studies that adopting data generated by Stable Diffusion to train a model can improve model generalization, we investigate the potential of utilizing such data to improve adversarial transferability. However, we find that existing input transformation-based attacks mainly utilize real data for augmentation, which may limit the adversarial transferability. In this paper, we propose a novel attack method named Stable Diffusion Attack Method (SDAM) which mixes the input image with the samples generated by Stable Diffusion for augmentation. In addition, we propose the fast version of SDAM to reduce the computational cost while maintaining high adversarial transferability. Extensive experimental results show that our method outperforms the state-of-the-art baselines by a clear margin. Moreover, our method is compatible with existing transfer-based attacks to further enhance the transferability of adversarial examples.

{
    \small
    \bibliographystyle{ieeenat_fullname}
    \bibliography{main}
}

\end{document}